\documentclass{article}

\usepackage[preprint]{neurips_2025}

\usepackage[utf8]{inputenc} 
\usepackage[T1]{fontenc}    
\usepackage{hyperref}       
\usepackage{url}            
\usepackage{booktabs}       
\usepackage{amsfonts}       
\usepackage{nicefrac}       
\usepackage{microtype}      
\usepackage{xcolor}         
\usepackage{cite}
\usepackage{graphicx}
\usepackage{amsmath}
\usepackage{wrapfig}
\usepackage{multirow}
\usepackage{enumitem}
\newcommand{\method}{\textbf{\textsc{PosS}}}

\title{\method:Position Specialist Generates Better Draft for Speculative Decoding}

\author{%
  Langlin Huang$^{1}$, Chengsong Huang$^{1}$, Jixuan Leng$^{2}$, Di huang$^{1}$, Jiaxin Huang$^{1}$ \\
  $^{1}$Washington University in St. Louis\\
  $^{2}$Carnegie Mellon University\\
  \texttt{h.langlin@wustl.edu},~~~\texttt{jiaxinh@wustl.edu} \\
}

\begin{document}

\maketitle

\begin{abstract}
Speculative decoding accelerates Large Language Model (LLM) inference by using a small draft model to predict multiple tokens, and a large target model to verify these tokens in parallel. 
Recent studies leverage the hidden state of the target model to enhance draft model prediction accuracy.
However, existing methods suffer from the degrading quality of draft token predictions at later positions, due to error accumulation in draft model generated features.
In this paper, we propose Position Specialists (\textbf{\method}), which consist of multiple position-specialized draft layers to generate tokens at assigned position(s). Position specialists greatly improve token acceptance rate at later positions per drafting round, as each specialist only needs to focus on handling a certain level of draft model feature deviation.
Experiment results on \text{Llama-3-8B-Instruct} and Llama-2-13B-chat across six datasets demonstrate that \textbf{\method} effectively improves over baselines on average acceptance length and speed-up ratio. Our codebase is available at \url{https://github.com/shrango/PosS}.
\end{abstract}

\section{Introduction}
Speculative decoding~\citep{Leviathan2022FastIF,Chen2023AcceleratingLL} is an effective approach to accelerate the autoregressive decoding of Large Language Models (LLMs) through a draft-then-verify framework. Specifically, it employs a lightweight draft model to generate candidate tokens autoregressively, which are then verified by the larger target model in parallel to determine accepted tokens from proposed draft tokens, thereby reducing overall decoding time.
The effectiveness of speculative decoding largely depends on the average acceptance length $\tau$ (accepted token counts per round) from the prediction depth $L$ (predicted token counts generated by the draft model per round).

Recent efforts~\citep{cai2024medusa, li2024eagle, li2024eagle2} in speculative decoding utilize the target model hidden states as input to enhance draft model prediction accuracy. EAGLE~\citep{li2024eagle, li2024eagle2} employs a one-layer Transformer as the draft model and trains it to predict the next token with features from the target model. However, EAGLE exhibits a training–inference discrepancy: during training, it predicts tokens using ground-truth features from the target model, whereas during autoregressive inference, ground-truth features at previous draft positions are unavailable. Instead, it must rely on features generated by the draft model, which deviate from the ground-truth. HASS~\citep{zhang2024learning} partially addresses this discrepancy by training the draft model to predict the next token with features from previous draft steps. However, both approaches suffer from relying on a single draft model to predict tokens at multiple positions in the draft sequence.

We hypothesize that \textbf{effective draft model should be position-specialized} within the prediction length $L$: early positions require accurate predictions with reliable target model features, while later positions must learn to mitigate the increasing levels of feature deviations. To evaluate the prediction quality across positions, we introduce the metric of position-wise acceptance rate (pos-acc) to measure the conditional probability of accepting the $i^\text{th}$ token given the acceptance of its preceding $(i-1)^\text{th}$ token. Our analysis reveals that both EAGLE and HASS suffer from rapidly degrading pos-acc beyond the first few predicted tokens. This confirms our hypothesis that a single draft model is limited by its generalization capability of various positions.

To address this challenge, we propose Position Specialists (\textbf{\method}), a novel framework that consists of multiple position-specialized draft layers, called as position specialists. Each position specialist is trained for predicting tokens at its assigned position(s), and only needs to handle an expected level of feature deviation at that position, thus enabling more accurate draft token predictions than a single draft model which needs to handle varying levels of feature deviation at different positions. 

We conduct extensive experiments on two model sizes (Llama-3-8B-Instruct and Llama-2-13B-chat) across six benchmark datasets, and demonstrate that \method~consistently outperforms baseline methods. \method~surpasses the strong baseline HASS on average acceptance length by up to 4.5\% (from 4.62 to 4.83) and on speed-up ratio by up to 5.7\% (from 2.97x to 3.14x). We also carry out comprehensive analysis and reveal that \textbf{the efficiency of \method~comes from reduced rounds of speculative generation}, as higher position-wise acceptance rate at deeper positions enables longer acceptance length $\tau$ per round.

Our primary contributions include:
\begin{itemize}[leftmargin=*]
\item We introduce position-wise acceptance rate (pos-acc) as a crucial metric for analyzing the draft quality of speculative decoding approaches.
\item We propose Position Specialists (\textbf{\method}), a novel framework that employs position-specialized layers to address the challenge of accumulated levels of feature deviation in draft predictions.
\item We conduct extensive experiments and analysis to demonstrate that \method~outperforms baseline methods on both average acceptance length and speed-up ratio.
\end{itemize}

\section{Preliminary}
\subsection{Speculative Decoding}
Speculative decoding harnesses the principle of speculative execution~\citep{Kung1979OnOM} to achieve reduced latency through increased parallelism. In this framework, 
a smaller, faster draft model $\theta_{D}$ works alongside a larger target language model $\theta_{T}$ that we aim 
to accelerate. 
The standard speculative decoding~\citep{Leviathan2022FastIF} process operates in three key phases. First, the draft model $\theta_{D}$ autoregressively generates a candidate sequence of length $L$. Next, the target model $\theta_{T}$ evaluates all $L$ draft tokens in parallel, computing their output distributions in a single forward pass. Finally, a specialized rejection sampling mechanism accepts 
tokens that align with the target distribution. This parallel evaluation significantly reduces inference latency compared to traditional token-by-token generation.

\subsection{Hidden State Assisted Speculative Decoding}
Recent research efforts~\citep{cai2024medusa, li2024eagle, li2024eagle2} discover the potential of the target model's hidden state. Instead of using a complete auxiliary model for drafting, researchers demonstrate that applying a few extra layers to process the last-layer hidden states of the target model, referred to as features, suffices for effective draft generation. Medusa~\citep{cai2024medusa} uses multiple language model heads to project a feature vector into different output spaces to predict several subsequent tokens simultaneously. 

EAGLE and EAGLE-2~\citep{li2024eagle, li2024eagle2} represents a significant breakthrough in speculative decoding through concatenating input embedding with feature vectors. It employs a one-layer Transformer as the draft model $\theta_{D}$ and reuses LM head of the target model for token prediction.
At generation step $t$, 
EAGLE-2's draft model $\theta_{D}$ predicts the next token $x_{t+1}$ based on context $x_{\le t}$ and features $f_{<t}$:

\begin{equation}\label{eq:eagle_inference}
P(x_{t+1}) = \text{Head}(\theta_{D}( [x_t; f^{(T)}_{t-1}], [x_{t-1}; f^{(T)}_{t-2}], \dots, [x_1; f^{(T)}_0]))
\end{equation}

\begin{figure}
    \centering
    \includegraphics[width=\linewidth]{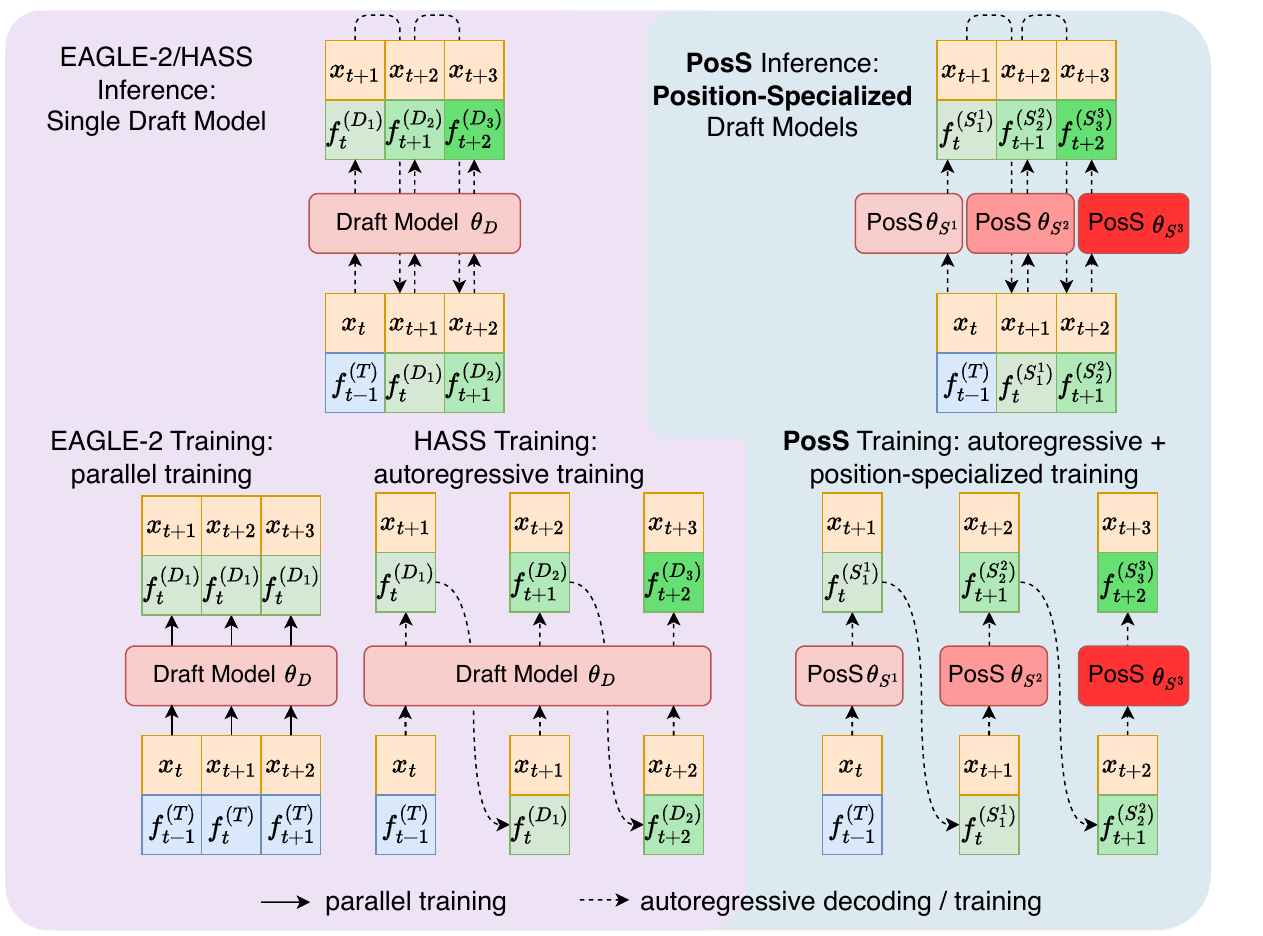}
    \caption{The inference and training stages of EAGLE-2, HASS, and our \method~method. The dashed lines represent autoregressive decoding or training, and the solid lines represent parallel training. The input concatenates context word embeddings $x$ and features from previous step~$f$.
    During \textbf{inference}, EAGLE-2 and HASS use a single draft model $\theta_{D}$ to generate features $f^{(D_i)}$ for each position $i$ recursively.
    For \textbf{draft model training}, EAGLE-2 uses the target model feature $f^{(T)}$ as input for training (teacher forcing), and HASS uses draft model predicted features $f^{(D_i)}$ to replace.
    Different from them, \textbf{\method}~introduces different position specialists $\theta_{S^{j}}$. During inference, the position-specialized draft models autoregressively generate features $f^{(S_i^j)}$, where position $i$ corresponds to the specialist $\theta_{S^{j}}$. At training stage, \method~applies position-specialized training: A specialist $\theta_{S^j}$ is trained on the $i^{\text{th}}$ position using the previous step specialist feature. 
    }
    \label{fig:method-comparison}
\end{figure}

Figure~\ref{fig:method-comparison} provides an example of EAGLE-2 at inference stage. $\theta_{D}$ autoregressively generates draft tokens $x_{t+1}, x_{t+2}, x_{t+3}$, where the subscripts represent the timesteps. Inputs are derived from different sources, denoted by superscripts: $f^{(T)}$ represents feature from the target model; $f^{(D_i)}$ represents feature from the $i^{th}$ draft step of the draft model $D$. $f^{(D)}$ is used instead of $f^{(T)}$ when the target model features are unavailable prior to the forward pass completion of subsequent tokens. Therefore, the prediction of the $k^{th}$ draft position is formulated as:

\begin{equation}\label{eq:hass_inference}
    P(x_{t+k}) = \text{Head}(\theta_{D}( [x_{t+k-1}; f^{(D_{k-1})}_{t+k-2}], \dots, [x_{t+1}; f^{(D_{1})}_{t}], [x_{t}; f^{(T)}_{t-1}], \dots, [x_1; f^{(T)}_0]))
\end{equation}

Specifically, Equation~\eqref{eq:hass_inference} degenerates to Equation~\eqref{eq:eagle_inference} when $k=1$. 

Although EAGLE-2 performs inference with Equation~\eqref{eq:hass_inference}, it is solely trained on Equation~\eqref{eq:eagle_inference}, known as ``teacher forcing''. 
This exhibits a fundamental training-inference discrepancy: $\theta_{D}$ needs to predict the subsequent tokens ($k>1$) with its own generated features during inference, but it never observes its own prediction errors during training, which impairs EAGLE-2's ability to effectively predict long draft sequences.

HASS~\citep{zhang2024learning} explicitly addresses the discrepancy through recursive feature alignment in draft model training, where $\theta_{D}$ is trained to predict subsequent tokens with its own generated features from earlier timesteps. Figure~\ref{fig:method-comparison} illustrates the training process of HASS. 
For the next $k>1$ token, HASS uses $f^{(D_{k-1})}$ generated by the draft model from the previous step to substitute for the target model feature.
Therefore, HASS is able to reuse each training token $L$ times, by considering it as the next $1^{\text{st}}$ to $L^{\text{th}}$ token in a sequence of draft predictions. By effectively training on the draft model generated features multiple times, HASS is able to improve the acceptance probabilities of tokens at later positions compared to EAGLE-2.

\section{Method}

In this section, we introduce our Position Specialist (\method) approach for speculative decoding. We first introduce the concept of position-wise acceptance rate to reveal the fundamental limitations in existing approaches in Section~\ref{sec:pos-acc}. We then propose our \method~with position specialized training in Section~\ref{sec:position_spec} to address the limitation. Finally, we analyze potential computational cost in real implementation in Section~\ref{sec:extra_cos}.

\subsection{Position-Wise Acceptance Rate}
\label{sec:pos-acc}
Previous speculative decoding frameworks rely heavily on the generalizability of a single draft layer for multi-position token generation. EAGLE-2 trains $\theta_{D}$ only on the immediate next position but expects it to generalize to subsequent positions at inference time. While HASS trains $\theta_{D}$ on both the immediate and later positions, it only uses one draft model to generalize across diverse feature sources and different token positions. Both EAGLE-2 and HASS use a single Transformer layer as the draft model, which inherently constrain the generalizability due to model capacity.

To demonstrate the generalization limitation of EAGLE-2 and HASS, we introduce \textbf{position-wise acceptance rate} (\textbf{pos-acc}), which measures the probability that a token at position $i$ is accepted given that its preceding token at position $i-1$ is accepted. Formally, the \textbf{pos-acc} at position $i$ is defined as:

\begin{equation}\label{eq:pos-acc}
    \text{\textbf{pos-acc}}_i = P(A_i \mid A_{i-1}) = \frac{P(A_{i-1} \cap A_i)}{P(A_{i-1})} = \frac{P( A_i)}{P(A_{i-1})}\ ,\ \ i > 1
\end{equation}

where $A_i$ denotes the event that the token at position $i$ is accepted during the verifying process. Notice that the target model acceptance follows a strict sequential dependency: if $x_i$ is accepted, its preceding tokens $x_{[0:i-1]}$ must also have been accepted, and therefore $A_i \subset A_{i-1}$.

We point out that higher \textbf{pos-acc} is crucial for achieving a higher acceptance length $\tau$ at each draft-verification round. For a draft sequence of length $L$, the probability of accepting all draft tokens up to position $k$ ($k \leq L$) is:
\begin{equation}\label{eq:acc-length}
P(A_1 \cap A_2 \cap \cdots \cap A_k) =
\begin{cases}
P(A_1) & \text{if } k=1\\
 P(A_1) \prod_{i=2}^k \text{\textbf{pos-acc}}_i  & \text{if } k>1
\end{cases}
\end{equation}

\begin{wrapfigure}{r}{0.45\textwidth}
  \centering
  \vspace{-1.0em}
  \includegraphics[width=0.45\textwidth]{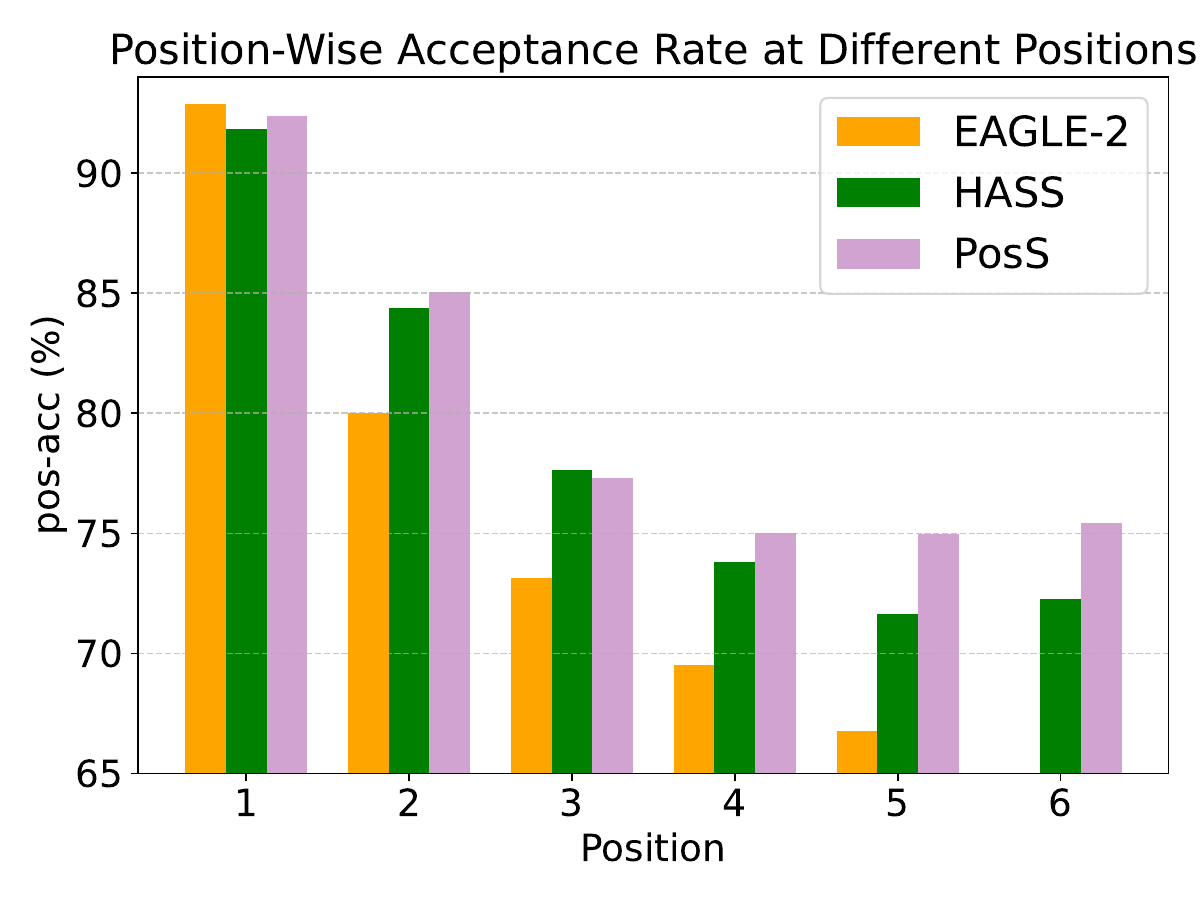}
  \vspace{-2.0em}
  \caption{Position-wise acceptance rate (\textbf{pos-acc}) of the $i^{th}$ token on MT\_Bench dataset by various speculative decoding methods. The \textbf{pos-acc} of EAGLE-2 and HASS decays fast as the draft sequence gets longer. Our proposed \method~method keeps a stable and higher \textbf{pos-acc} even at the deepest position (draft model prediction depth $L=6$).}
  \vspace{-3.5em}
  \label{fig:conditional_acc_ratio}
\end{wrapfigure}

This chain rule decomposition reveals that the overall acceptance length depends on the multiplication of \textbf{pos-acc}, and is particularly sensitive to degradation in any single position. Notably, token prediction inherently becomes more challenging at later positions due to the accumulation of prediction errors and the increasing uncertainty in longer draft positions.

In Figure~\ref{fig:conditional_acc_ratio}, we demonstrate the empirical \textbf{pos-acc} of EAGLE-2 and HASS.
EAGLE-2's \textbf{pos-acc} deteriorates rapidly beyond position $k=1$. This is because the draft model of EAGLE-2 is solely trained on predicting the next immediate token.
HASS is able to maintain relatively higher \textbf{pos-acc} because its draft model is trained on multiple subsequent positions. However, since its single draft model needs to balance between multiple positions,
the \textbf{pos-acc} drops by about 1 percent at position $k=1$, which critically impairs the overall acceptance length due to the multiplicative nature of the acceptance probability in Equation~\eqref{eq:acc-length}.

\subsection{Position Specialists Improve Position-Wise Acceptance Rate}

\label{sec:position_spec}

To address the aforementioned limitation, we introduce Position Specialists (\method)
to preserve early-position acceptance rate while enhancing later position predictions. 
\method~consists of multiple position-specialized draft layers, called position specialists. Each specialist is trained for certain position(s) and generates draft tokens at its assigned position(s). The number of positions that a specialist is assigned to can be pre-defined as $n$, and \method-$n$ means each specialist is responsible for $n$ positions. Figure~\ref{fig:method-comparison} exhibits the training and inference of \method-$1$. In the example, there are 3 position specialists $\{\text{\method}^i\}_{i=1}^3$, with each assigned to predict the draft token $x_{t+i}$. During training, each specialist $\text{\method}^i$ learns to predict using input feature of draft model at previous step $\text{\method}^{i-1}$.

\begin{figure}
    \centering
    \includegraphics[width=\linewidth]{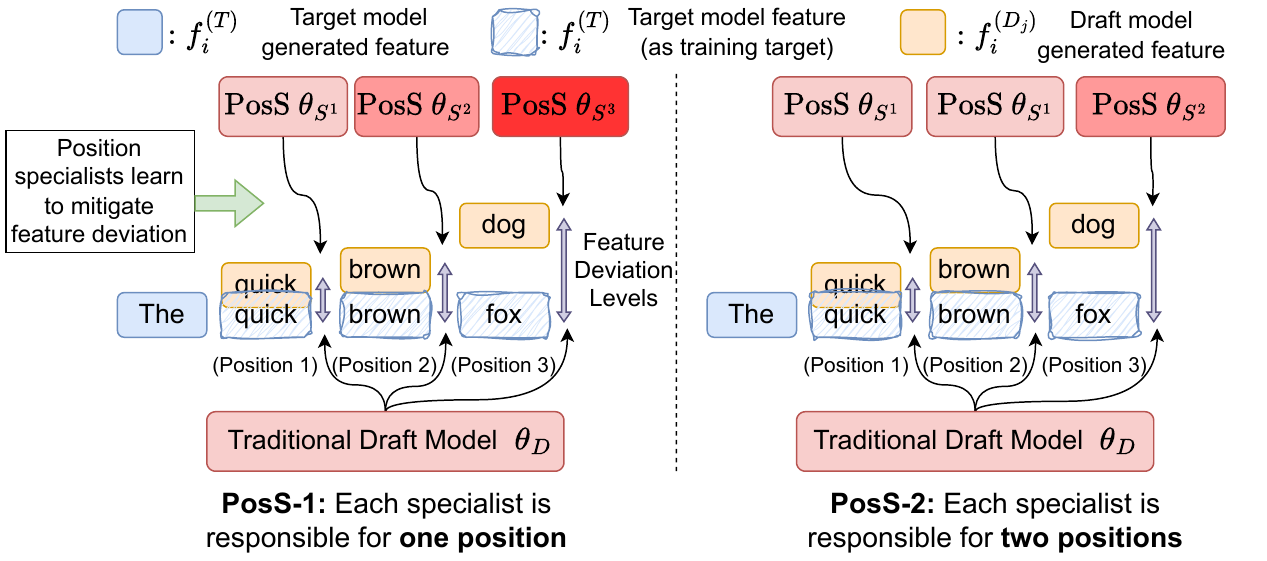}
    \caption{This figure shows a comparison of hidden state (feature) training between \method~and previous work. The draft model generated feature inevitably deviates from the target model, and the deviation level increases with error propagation in the draft model(s). In previous work, one draft model learns to mitigate 
    varying levels of feature gaps. In \method~method, position specialists can focus on mitigating expected levels of feature deviation.
    }
    \label{fig:hidden-state-training-diff}
\end{figure}

Figure~\ref{fig:hidden-state-training-diff} illustrates the training of \method~to explain why position specialists are better than a single draft model. Despite being trained, the draft model inevitably generates hidden states (features) that deviate from the target model. The deviation  $||f^{(D)}_i - f^{(T)}_i||$ increases with position $i$ through error propagation in the autoregressive draft process, as visualized by the purple arrow in Figure~\ref{fig:hidden-state-training-diff}. While previous approaches like HASS train a single draft model to handle varying levels of input noise -- a challenging task for a single-layer architecture, our \method~method employs position-specific specialists to handle the narrow and expected levels of feature deviation
, enabling more accurate draft sequence prediction by decomposing it into position-specific subtasks.   

We introduce the loss of \method-n as follows. A token-level loss is applied to guide draft model to generate the same token as target model does. Given verified context length $t$, the token-level loss on the $i^{th} (i\ge0)$ position of the draft sequence is 

\vspace{-1.5em}
\begin{equation}
    L_{\text{token}}=\mathrm{Cross\_Entropy}(P(x_{t+i+1}), \hat{P}(x_{t+i+1}))
\end{equation}
\vspace{-1.0em}
where the training target $\hat{P}(x_{t+i+1})$ and the training variable $P(x_{t+i+1})$ are given by:

\begin{equation}
    \hat{P}(x_{t+i+1})=\text{Head}(f^{(T)}_{t+i})
\end{equation}
and
\begin{equation}
    P(x_{t+i+1}) = \text{Head}(f^{(S^{\left\lceil (i+1)/n \right\rceil})}_{t+i})
\end{equation}
Here, $f^{(T)}_{t+i}$ and $f^{(S^{\left\lceil (i+1)/n \right\rceil})}_{t+i}$ represent the target feature and the specialist feature, where $\left\lceil (i+1)/n \right\rceil$ determines which specialist is responsible for position $i$. For example, with $n=2$, specialist $S^1$ handles positions 0 and 1, specialist $S^2$ handles positions 2 and 3, and so forth. The target and specialist features are computed as
\begin{equation}
    f^{(T)}_{t+i} = \theta_{T}([x_{t+i}; f^{(T)}_{t+i-1}], [x_{t+i-1}; f^{(T)}_{t+i-2}], \dots, [x_1; f^{(T)}_0])
\end{equation}
and
\begin{equation}
    f^{(S^{\left\lceil (i+1)/n \right\rceil})}_{t+i} = \theta_{(S^{\left\lceil i/n \right\rceil})}( [x_{t+i}; f^{(S^{\left\lceil i/n \right\rceil})}_{t+i-1}], \dots, [x_{t+1}; f^{(S^{1})}_{t}], [x_{t}; f^{(T)}_{t-1}], \dots, [x_1; f^{(T)}_0])
\end{equation}

Besides token-level loss, we also employ feature-level loss. This loss trains the specialist models to generate features that are close to the target features, which is given by:
\begin{equation}
    L_{\text{feature}} = \mathrm{Smooth~L1}(f^{(T)}_{t+i}, f^{(S^{\left\lceil (i+1)/n \right\rceil})}_{t+i}) 
\end{equation}

Since the specialist draft model takes in deviated features from preceding draft models during training, it effectively learns to mitigate the accumulated feature deviation at inference time.

We also adopt the Top-K distillation loss $L_{\mathrm{Top-K}}$, proposed by HASS~\citep{zhang2024learning}, to ensure a fair comparison with it. The loss is defined as:

\begin{equation}
    L_{\mathrm{Top-K}}=\mathrm{Cross\_Entropy}_{\mathrm{Top-K}}(P(x_{t+i+1}),\hat{P}(x_{t+i+1}))
\end{equation}

which calculates cross-entropy loss only on $K$ tokens of the highest probabilities from $\hat{P}(x_{t+i+1})$.

The overall loss is $L_{\text{total}}=L_{\text{feature}}+w\times L_{\text{token}}$
$+L_{\mathrm{Top-K}}$
, where we follow previous works~\citep{li2024eagle,li2024eagle2,zhang2024learning} and set weight $w=0.1, K=10$.

\subsection{Computational Overhead of Position Specialists}
\label{sec:extra_cos}
While \method~generates drafts closer to the target model and achieves longer acceptance length, we point out two types of additional computation overhead that \method~introduces.

First, the GPU memory usage increases linearly with the number of position specialists. Fortunately, this additional cost is negligible compared to the target model size since each specialist costs only one transformer layer (around 218M parameters per specialist for an 8B target model).

Second, the switching of position specialists brings a little extra latency.
Although each \method~specialist use the same structure with the single draft model of EAGLE-2 -- theoretically implying equivalent computation time per draft phase -- practical implementation of position specialists costs slightly more computation overhead for two reasons: (1) Non-shared KV cache across layers: Each position specialist computes key-value cache for draft tokens generated by its preceding specialist in addition to previously verified tokens.
(2) Parameter switching overhead: Frequent parameter switching between specialists may introduce additional latency due to hardware-level parameter loading.

We study the effect of extra computation overhead brought by \method~through a comprehensive empirical analysis in Section~\ref{sec:analysis-tradeoff}. We demonstrate that \method~only brings minimal overhead compared to the overall computation time, and this overhead is largely outweighed by the increased average acceptance length, which reduces the overall drafting rounds needed.

\section{Experiment}

\subsection{Experiment Setup}
\label{sec:experiment-setups}
\paragraph{Metrics.} 
We evaluate the performance of our approach using two key metrics: speed-up ratio and average acceptance length. 
\begin{itemize}[leftmargin=*]
\item \textbf{Speed-up Ratio}: The speed-up ratio measures the improvement in generation efficiency compared to the vanilla target model decoding, calculated as the ratio between throughputs (tokens generated per second) of a speculative decoding approach to that of the target model autoregressive decoding. A higher speed-up ratio indicates better performance. 
\item \textbf{Average Acceptance Length $\tau$}: The average acceptance length represents the mean number of tokens accepted in each round of $L$ drafting positions (denoted as prediction length). It reflects how effectively the draft model can predict longer sequences that match the target model output. Longer acceptance lengths generally correlate with improved efficiency as they reduce the number of draft iterations needed.
\end{itemize}

\paragraph{Datasets.}
We conduct comprehensive experiments on six datasets, following EAGLE-2~\citep{li2024eagle}. This includes MT-Bench~\citep{zheng2023judgingllmasajudgemtbenchchatbot} for multi-turn conversation, Alpaca~\citep{alpaca} for instruction following,  GSM8K~\citep{cobbe2021trainingverifierssolvemath} for mathematical reasoning, Natural Questions~\citep{kwiatkowski-etal-2019-natural} for question answering, CNN/Daily Mail (shortened to CNN/DM)~\citep{nallapati2016abstractivetextsummarizationusing} for summarization, and HumanEval~\citep{chen2021evaluatinglargelanguagemodels} for code generation. 

\paragraph{Target Models.}
We evaluate our method on two model sizes: Llama-3-8B-Instruct and Llama-2-13B-chat. This allows us to evaluate how our approach performs across model sizes. Llama-3-8B-Instruct serves as our primary model for ablation studies and detailed analysis, while Llama-2-13B demonstrates the scalability of our method to larger models.

\paragraph{Implementations.}

Our implementation is built upon the open-source repositories of EAGLE-2 and HASS. 
We experiment with EAGLE-2, HASS, and our method with configurations of \method-1, 2, 3, where \method-3 adds the least extra layers and computation overhead. 
The training configurations are mostly aligned with HASS and are detailedly introduced in Appendix~\ref{appendix:configuration}.
During inference, tree-drafting~\citep{chen2024sequoia} strategy is applied to generate multiple draft paths in one draft phase, where the \textit{width}, \textit{depth}, and \textit{total tokens} are key controlling factors. We set the draft tree \textit{width} to 10 and the number of draft \textit{total tokens} to 60 for all experiments. We choose the draft tree \textit{depth} that leads to the best performance. Table~\ref{tab:hyper-speedup-llama3} and \ref{tab:hyper-speedup} suggest that the 8B target model setting achieves the best performance at \textit{depth}=6, and the 13B target model reaches the best performance at \textit{depth}=7. All experiments are conducted on NVIDIA A100 GPUs with 80GB of memory. We repeat all experiments 3 times and select the fastest one among each method for speed-up ratio calculation. This mitigates the random disturbance of the server and better reflects the real throughputs and speed-up ratio.
\vspace{-1.0em}
\section{Results}
\label{sec:results}
We introduce the main results in this section. Table~\ref{tab:acc} presents the average acceptance lengths of different models. Table~\ref{tab:speedup} presents the speed-up ratio of these models.

Our methods achieve the highest overall average acceptance length under different sampling temperatures, demonstrating the effectiveness of position specialists in making accurate draft predictions.
When L3 8B serves as the target model, \method~achieves consistently higher speed-up ratio over the baselines. When L2 13B is the target model and generates stronger feature representations, \method~is less advantageous, but \method-3 still achieves the highest speed-up ratio.

\begin{table}[h]
\caption{Average acceptance length $\tau$ of all methods. L3 8B represents Llama-3-8B-Instruct, L2 13B represents Llama-2-13B-Chat. 
}
\label{tab:acc}
\centering
\resizebox{\textwidth}{!}{
\begin{tabular}{ccccccccc}
\toprule
\multicolumn{9}{c}{Temperature=0}                                                                                                 \\ \hline
\multicolumn{1}{c|}{Model}                   & Method & MT-Bench & Alpaca & GSM8K & Natural Questions & CNN/DM & HumanEval & Avg. \\ \hline
\multicolumn{1}{c|}{\multirow{5}{*}{L3 8B}}  & EAGLE-2  & 4.11     & 4.32   & 4.25  & 3.38              & 3.61   & 4.70      & 4.06 \\
\multicolumn{1}{c|}{}                        & HASS   & 4.42     & 4.62   & 4.61  & 3.54              & 3.92   & 5.20      & 4.39 \\
\multicolumn{1}{c|}{}                        & \method-1 (ours) & \textbf{4.54}     & 4.78   & 4.82  & \textbf{3.65}              & \textbf{4.06}   & 5.39      & 4.54 \\
\multicolumn{1}{c|}{}                        & \method-2 (ours) & \textbf{4.54}     & \textbf{4.83}   & \textbf{4.83}  & 3.63              & \textbf{4.06}   & 5.40      & \textbf{4.55} \\
\multicolumn{1}{c|}{}                        & \method-3 (ours) & 4.52     & 4.82   & 4.81  & 3.64              & 4.05   & \textbf{5.41}      & 4.54 \\ \hline
\multicolumn{1}{c|}{\multirow{5}{*}{L2 13B}} & EAGLE-2  &   4.86   &  4.64  &  5.01  &    4.15    &   4.30   &   5.78   &  4.79  \\
\multicolumn{1}{c|}{}                        & HASS   &    \textbf{5.40}    &   \textbf{5.31}   &  5.47   &     4.55        &  4.71  &    6.47   &  5.32 \\
\multicolumn{1}{c|}{}                        & \method-1 (ours)   & \textbf{5.40}        & 5.23      & 5.60     & 4.57                 & 4.78      & 6.48         & 5.34    \\
\multicolumn{1}{c|}{}                        & \method-2 (ours) & 5.39        & 5.27      & 5.58     & 4.59                 & 4.79      & \textbf{6.51}         & 5.36    \\
\multicolumn{1}{c|}{}                        & \method-3 (ours) &   \textbf{5.40}     &  5.28   &  \textbf{5.62}  &    \textbf{4.60}   &  \textbf{4.80}   &    \textbf{6.51}   &  \textbf{5.37}  \\ \hline
\multicolumn{9}{c}{Temperature=1}                                                                                                 \\ \hline
\multicolumn{1}{c|}{\multirow{5}{*}{L3 8B}}  & EAGLE-2  & 4.11        & 4.32      & 4.25     & 3.38                 & 3.61      & 4.70        & 4.06    \\
\multicolumn{1}{c|}{}                        & HASS    & 4.01        & 4.39      & 4.49     & 3.40                 & 3.65      & 5.00         & 4.16    \\
\multicolumn{1}{c|}{}                        & \method-1 (ours) &  \textbf{4.18}       &   4.46    &   4.63   &  3.41          &  \textbf{3.78}   &   \textbf{5.20}   &  \textbf{4.28} \\
\multicolumn{1}{c|}{}                        & \method-2 (ours) & 4.17        & \textbf{4.47}      & 4.66     & \textbf{3.44}                 & 3.77      & 5.13         & 4.27    \\
\multicolumn{1}{c|}{}                        & \method-3 (ours) & 4.13        & 4.46      & \textbf{4.67}     & 3.37                 & 3.76      & 5.12         & 4.25    \\ \hline
\multicolumn{1}{c|}{\multirow{5}{*}{L2 13B}} & EAGLE-2  & 4.69        & 4.44      & 4.82     & 4.12                 & 4.25      & 5.54         & 4.64   \\
\multicolumn{1}{c|}{}                        & HASS   & 5.14        & \textbf{5.21}      & 5.32     & 4.46                 & 4.54      & 6.16         & 5.14    \\
\multicolumn{1}{c|}{}                        & \method-1 (ours)   &  \textbf{5.31}      &   5.19    &   \textbf{5.57}   &     4.45     &  4.72  &   6.20   &   \textbf{5.24} \\
\multicolumn{1}{c|}{}                        & \method-2 (ours) & 5.17        & 5.05      & 5.44     & 4.41                 & 4.67      & 6.22         & 5.16    \\
\multicolumn{1}{c|}{}                        & \method-3 (ours) & 5.20        & 5.18      & 5.49     & \textbf{4.50}                 & \textbf{4.73}      & \textbf{6.30}         & 5.23   \\
\bottomrule
\end{tabular}
}
\end{table}

\begin{table}[h]
\caption{Speed-up ratios of all methods. L3 8B represents Llama-3-8B-Instruct, L2 13B represents Llama-2-13B-Chat. 
}
\label{tab:speedup}
\centering
\resizebox{\textwidth}{!}{
\begin{tabular}{ccccccccc}
\toprule
\multicolumn{9}{c}{Temperature=0}                                                                                                 \\ \hline
\multicolumn{1}{c|}{Model}                   & Method & MT-Bench & Alpaca & GSM8K & Natural Questions & CNN/DM & HumanEval & Avg. \\ \hline
\multicolumn{1}{c|}{\multirow{5}{*}{L3 8B}}  & EAGLE-2  & 2.77x     & 2.79x   & 2.87x  & 2.29x    & 2.27x   & 3.08x      & 2.68x \\
\multicolumn{1}{c|}{}                        & HASS   & 2.94x     & 2.97x   & 3.11x  & 2.38x      & 2.47x   & 3.48x      & 2.89x \\
\multicolumn{1}{c|}{}                        & \method-1 (ours) & 2.96x     & 3.00x   & 3.19x  & \textbf{2.49x}     & 2.50x   & 3.52x      & 2.94x \\
\multicolumn{1}{c|}{}                        & \method-2 (ours) & \textbf{2.99x}     & \textbf{3.14x}   & \textbf{3.25x}  & 2.45x              & \textbf{2.52x}   & 3.52x      & \textbf{2.98x} \\
\multicolumn{1}{c|}{}                        & \method-3 (ours) & 2.96x     & 3.10x   & 3.17x  & 2.45x              & 2.50x   & \textbf{3.53x}      & 2.95x \\ \hline
\multicolumn{1}{c|}{\multirow{5}{*}{L2 13B}} & EAGLE-2  & 2.99x     & 2.95x   & 3.23x  & 2.71x              & 2.49x   & 3.71x      & 3.01x \\
\multicolumn{1}{c|}{}                        & HASS   & \textbf{3.28x}     & \textbf{3.34x}   & 3.52x  & \textbf{2.96x}              & 2.72x   & \textbf{4.15x}      & 3.33x \\
\multicolumn{1}{c|}{}                        & \method-1 (ours) & 3.16x        & 3.18x      & 3.47x     & 2.86x   & 2.63x      & 3.99x         & 3.21x    \\
\multicolumn{1}{c|}{}                        & \method-2 (ours) & 3.22x        & 3.26x      & 3.54x     & 2.93x                 & 2.72x      & 4.11x         & 3.30x    \\
\multicolumn{1}{c|}{}                        & \method-3 (ours) & \textbf{3.28x}     & 3.32x   & \textbf{3.59x}  & \textbf{2.96x}             & \textbf{2.74x}   & 4.12x      & \textbf{3.34x} \\ \hline
\multicolumn{9}{c}{Temperature=1}                                                                                                 \\ \hline
\multicolumn{1}{c|}{\multirow{5}{*}{L3 8B}}  & EAGLE-2  & 2.67x        & 2.55x      & 2.09x     & 2.02x                 & 2.80x      & 2.47x         & 2.43x    \\
\multicolumn{1}{c|}{}                        & HASS   & 2.77x        & 2.79x      & 2.14x     & 2.09x                 & 3.03x      & 2.56x         & 2.56x    \\
\multicolumn{1}{c|}{}                        & \method-1 (ours) & 2.79x     & 2.78x   & 2.13x  & \textbf{2.19x}   & \textbf{3.18x}   & \textbf{2.67x}      & \textbf{2.62x} \\
\multicolumn{1}{c|}{}                        & \method-2 (ours) & \textbf{2.81x}     & 2.79x   & \textbf{2.17x}  & 2.18x              & 3.10x   & 2.55x      & 2.60x \\
\multicolumn{1}{c|}{}                        & \method-3 (ours) & 2.71x     & \textbf{2.86x}   & 2.12x  & 2.18x              & 3.11x   & 2.58x      & 2.59x \\ \hline
\multicolumn{1}{c|}{\multirow{5}{*}{L2 13B}} & EAGLE-2  & 2.95x        & 2.88x      & 3.13x     & 2.76x                 & 2.51x      & 3.48x         & 2.95x \\
\multicolumn{1}{c|}{}                        & HASS   & \textbf{3.22x}        & \textbf{3.30x}     & 3.46x     & \textbf{2.97x}                 & 2.67x      & 3.89x         & 3.25x    \\
\multicolumn{1}{c|}{}                        & \method-1 (ours)        & 3.16x     & 3.19x     & 3.44x                 & 2.86x      & 2.65x         & 3.75x  & 3.17x     \\
\multicolumn{1}{c|}{}                        & \method-2 (ours) & 3.16x        & 3.17x      & 3.42x     & 2.91x                 & 2.64x      & 3.85x         & 3.19x    \\
\multicolumn{1}{c|}{}                        & \method-3 (ours) & \textbf{3.22x}       & 3.23x      & \textbf{3.49x}     & 2.96x                 & \textbf{2.73x}     & \textbf{3.92x}         & \textbf{3.26x}   \\
\bottomrule
\end{tabular}
}
\end{table}

\section{Analysis}\label{sec:analysis}

\subsection{Position-Wise Acceptance Rate}
\begin{wrapfigure}{r}{0.5\textwidth}
 \vspace{-2.0em}
  \centering
  \includegraphics[width=0.5\textwidth]{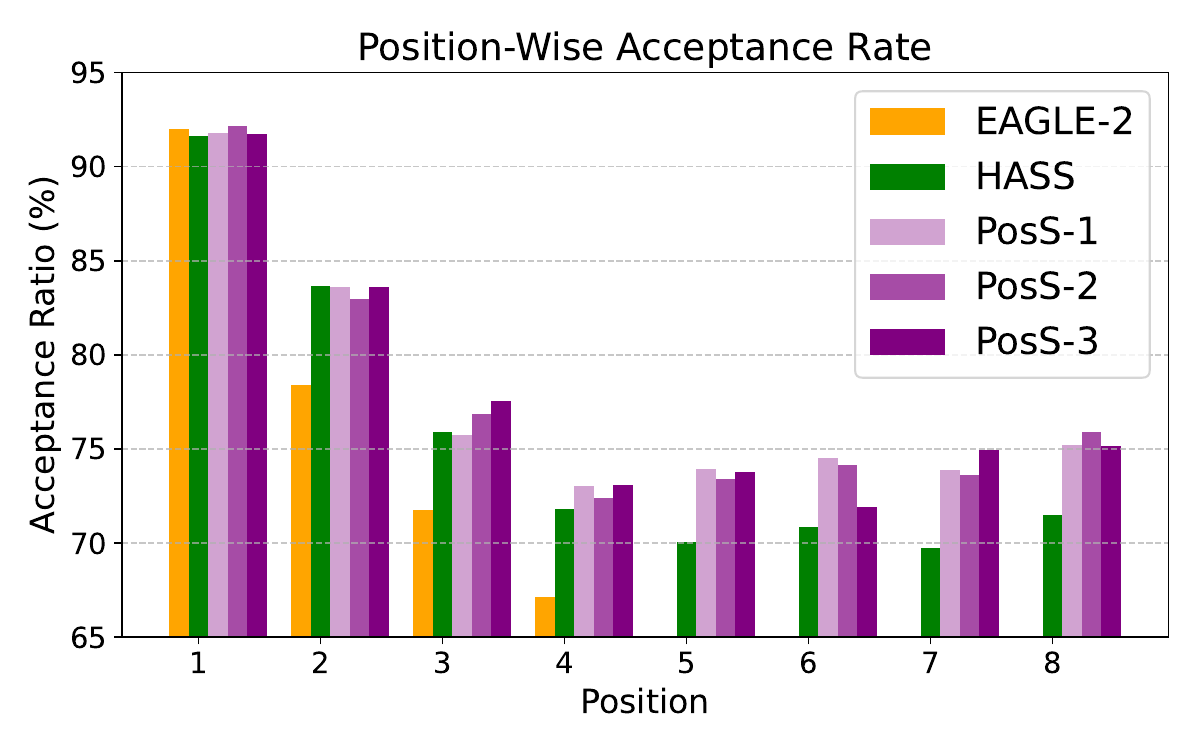}
  \vspace{-2.0em}
  \caption{The position-wise acceptance rate of EAGLE-2, HASS, and variants of \method. Experiments are conducted on MT-Bench dataset, with base model Llama-3-8B-Instruct and draft depth=8. Three variations of our method maintain a relatively higher pos-acc even at the $8^{th}$ position.}
  \label{fig:analysis-pos-acc}
  \vspace{-1.0em}
\end{wrapfigure}

In Section~\ref{sec:pos-acc}, we introduce the metric \textbf{position-wise acceptance rate} (\textbf{pos-acc}) to reflect the acceptance rate of a specific position, which largely affects the overall acceptance length. Here we demonstrate that \textbf{\method~largely improves pos-acc by mitigating the feature deviation at each position}. In Figure~\ref{fig:analysis-pos-acc}, we show the pos-acc with a draft depth of 8 on different models. 
EAGLE-2, with the least position generalization ability, has pos-acc lower than 65\% from the $5^{th}$ position on. HASS can only maintain adequate pos-acc at the first four positions, after which performance degrades significantly due to one single draft model. In contrast, all variants of our \method~method maintain substantially higher pos-acc until the last position. This demonstrates the effectiveness of \method~in mitigating position deviation and making accurate predictions.

\subsection{Computational 
Efficiency Tradeoffs for Extra Position Specialists}
\label{sec:analysis-tradeoff}
\begin{figure}
    \centering
    \includegraphics[width=\linewidth]{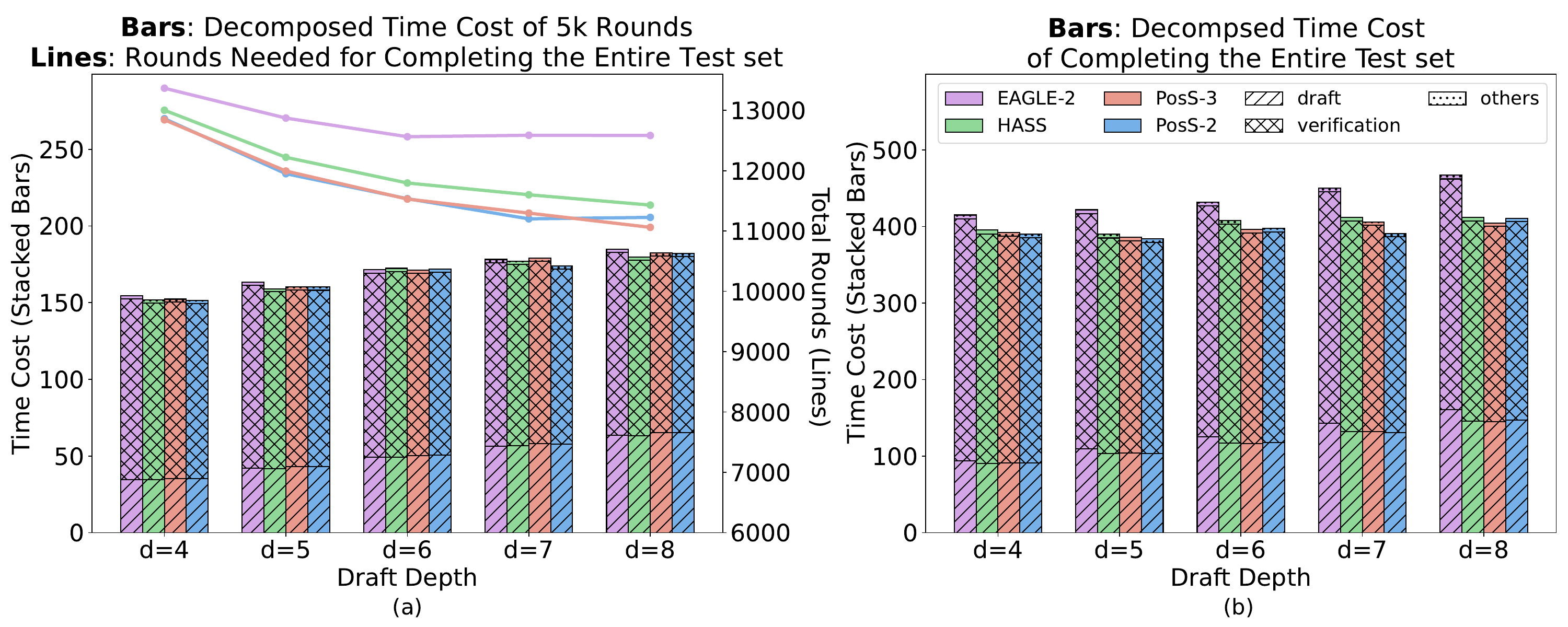}
    \caption{Computation time of different phases on MT-Bench dataset on different models across varying draft depths. The bar plots present the decomposition of time spent on each phase of speculative decoding, where subfigure (a) measures the time spent on 5k rounds and subfigure (b) measures the time to complete an entire test set. The line plot presents the number of rounds needed to complete a dataset. The lower the metrics are, the better the method is.}
    \label{fig:decomposition}
\end{figure}

In Section~\ref{sec:extra_cos}, we analyze the computational overhead of position specialists. Here we conduct a comprehensive analysis of computational costs and efficiency benefits brought by position specialists.
Each complete round of speculative generation involves two primary phases: the \textbf{draft phase} and the \textbf{verification phase}. 
In this experiment, we quantitatively analyze the time cost through three key metrics: (1) per-round computation time, (2) total round counts for test set generation, and (3) total time cost for test set generation.
We demonstrate a comprehensive analysis in Figure~\ref{fig:decomposition} and present the following noteworthy observations.

\paragraph{Position specialists bring minimal overhead to overall computation time.}
As discussed in Section~\ref{sec:extra_cos}, position specialists cost additional computational overhead.
We present in Figure~\ref{fig:decomposition}(a) the sum of per-round computation time over 5,000 rounds across varying draft depths (bar chart), decomposed into draft phases and verification phases. Empirical results show that \method~has lower per-round time than EAGLE-2. Compared to HASS, \method~only brings a negligible fraction of time in the draft phase with positional specialists, and keeps similar verification phase costs.

\paragraph{\method~achieves lowest overall computation time with reduced round counts.}
The line chart in Figure~\ref{fig:decomposition}(a) illustrates the total round counts needed for test set generation.
\method-2 and \method-3 consistently require fewer rounds than baseline methods, benefitting from accurate draft token prediction from position specialists.
The total time cost for decoding is primarily determined by both the \textbf{per-round time cost} and the \textbf{total round counts}. 
As shown in Figure~\ref{fig:decomposition}(b), \method-2 and \method-3 achieve lower overall time costs compared to EAGLE-2 and HASS. This confirms that the efficiency gains from \textbf{reduced round costs substantially outweigh the marginal per-round overhead} brought by position specialists.

\subsection{Ablation Study on Draft Model Prediction Depth}
\begin{wrapfigure}{r}{0.5\textwidth}
\vspace{-2.0em}
  \centering
\includegraphics[width=0.5\textwidth]{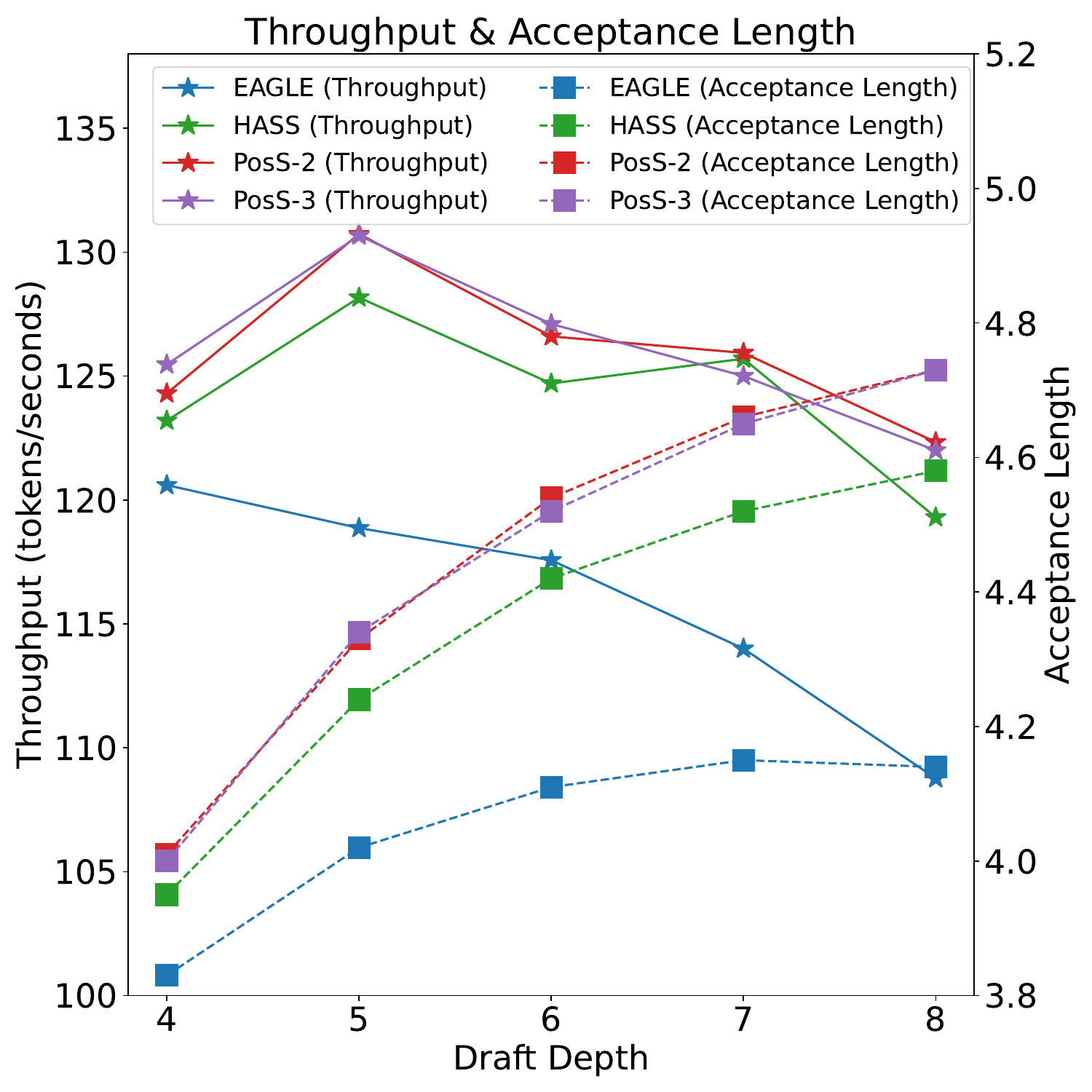}
  \caption{The throughput and average acceptance length of 4 models on different draft depths. The experiments are conducted on MT-Bench dataset. The acceptance length consistently increases as depth rises, while the throughput peaks at depth=5.}
  \label{fig:vary-depth}
  \vspace{-5.0em}
\end{wrapfigure}

Figure~\ref{fig:vary-depth} presents the throughput and average acceptance length under different draft depths. The average acceptance length $\tau$ increases with the draft depth consistently, but the improvement diminishes at higher depth. 
The diminishing improvement, along with the linearly increasing time cost of draft depth, creates an optimal point for throughput.
We empirically demonstrate that the throughput peaks at draft depth = 5 on the MT-Bench dataset for most models. We extend the ablation experiment on all six datasets in Appendix~\ref{appendix:varing-hyperparameters}. Results in Table~\ref{tab:speedup} demonstrates that \method~methods achieve speed-up ratio peaks at the $8^{th}$ position while HASS peaks at only the $7^{th}$ position.

\vspace{-1.0em}
\section{Related Work}
\subsection{Linear Speculative Decoding}
Early works~\citep{xia2022speculative} introduce the fundamental concept of using a draft model to predict multiple tokens in parallel. This is followed by various improvements in linear speculative decoding, including adaptive calibration techniques~\citep{gautam2025token}, dynamic candidate length adjustment~\citep{huang2024specdec++}, and methods to optimize the latency-throughput tradeoff~\citep{sadhukhan2024magicdec}. Recent advances focus on multi-token prediction~\citep{gloeckle2024better}, efficient multi-sampling~\citep{ni2024ems}, and token recycling~\citep{luo2024turning}. Some also explore parallel decoding strategies with adaptive n-gram techniques~\citep{ou2024lossless,wu2024parallel,liu2024parallel,wei2024fast}.

\subsection{Tree Speculative Decoding}
Tree-based speculative decoding has advanced through several key works. GRIFFIN~\citep{hu2025griffin} and Sequoia~\citep{chen2024sequoia} enhance token alignment methods, while SpecInfer~\citep{miao2024specinfer}improves sampling techniques. Other notable approaches include dynamic tree pruning~\citep{zhong2024propd}, early exit mechanisms~\citep{elhoushi2024layerskip}, and hierarchical method~\citep{sun2024triforce}.

\subsection{Efficient Inference}
Recent works apply other methods to improve the inference speed. Judge Decoding~\citep{bachmann2025judge} uses a small judge model to evaluate parallel reasoning paths, while SpecReason~\citep{pan2025specreason} and Speculative Thinking~\citep{yang2025speculative} leverage speculative computation for faster inference. Other efficient reasoning techniques include efficient chain-of-thought methods~\citep{wang2025efficient, huang2025efficient}, in-context learning methods~\citep{huang2024divide}, non-myopic generation~\citep{ma2024non} and system-level infra~\citep{10620276}.

\section{Conclusion}
This paper proposes \method, a draft model consisting of several position specialists. This method mitigates feature deviation between the draft and target models, and reduces the deviation accumulation across draft positions. Experiments show that \method~maintains a high position-wise acceptance rate at large positions, achieving a larger acceptance length and faster generation speed than other methods.

\bibliographystyle{unsrt}
\bibliography{references}

\newpage

\appendix

\section{Limitations}
\label{appendix:limitation}
\method~requires slightly more memory to deploy extra position-specialized layers, which is quantitatively discussed in Appendix~\ref{appendix:memory-usage}. Besides, these layers cannot share their key-value cache.
However, such computational overhead can be outweighed by the benefits from more accurate drafts generated by position specialists, as discussed in Section~\ref{sec:analysis-tradeoff}.

\section{Training Configurations}
\label{appendix:configuration}
In this work, all draft models are trained on the ShareGPT dataset and its distilled version generated by the corresponding target model. The former provides the token-level training objective, and the latter provides the feature-level training objective. 

When Llama-3-8B-Instruct serves as the target model, we notice that the open-sourced checkpoints of EAGLE-2 and HASS are inconsistent in structure\footnote{In the configuration file of EAGLE-2 model, \url{https://huggingface.co/yuhuili/EAGLE-LLaMA3-Instruct-8B/blob/main/config.json}, the "\textit{bias}" is false. However, this is true in the configuration file of HASS, \url{https://huggingface.co/HArmonizedSS/HASS-LLaMA3-Instruct-8B/blob/main/config.json}.}. To ensure the fairness of the comparison, we reproduce both methods with the model structure of HASS. While EAGLE-2 is trained for 20 epochs, HASS is trained for 40 epochs. To save the training time, we start training our method~\method~from the reproduced EAGLE-2 model for another 20 epochs to fairly compare with HASS.

When Llama-2-13B-Chat serves as the target model, we use the open-sourced checkpoints of EAGLE-2 and HASS. Similarly, we train~\method~from the EAGLE-2 model checkpoint for another 20 epochs.

During the training of \method~models, we keep most configurations the same with HASS~\citep{zhang2024learning}, including the loss-related hyperparameters and the learning rate.

\section{Different Drafting Hyperparameters}
\label{appendix:varing-hyperparameters}
Many factors influence the average acceptance length and speed-up ratio. Besides the prediction accuracy of draft models and computational overhead, the structure of draft trees also matters. We examine two key hyperparameters that affect the performance: depth and total tokens.

We take Llama-2-13B-chat as the base model, and conduct experiments with depths from 6 to 9, and total tokens selected from \{60, 80\}. We evaluate the models on all six datasets and take the average of them. Table~\ref{tab:hyper-acc} presents the average acceptance length, and Table~\ref{tab:hyper-speedup} presents the speed-up ratio.

Interestingly, despite the consistent rise of average acceptance length as the number of total tokens increases from 60 to 80, the speed-up ratio shows a sharp drop. This indicates the target model takes significantly more time to verify. This phenomenon results from the inner structure of the A100 GPU device that we use for experiments, which is also observed by OPT-Tree~\citep{wang-etal-2025-opt}.

\begin{table}[h]
\caption{Average acceptance length under different hyperparameters. Experiments use Llama-3-8B-Instruct as the base model. We average the results on all six datasets. The largest average acceptance length within each column is bolded.
}
\label{tab:hyper-acc-llama3}
\centering
\resizebox{\textwidth}{!}{
\begin{tabular}{cc|cccccccc}
\toprule
\multirow{2}{*}{Temperature} & Depth        & \multicolumn{2}{c}{6} & \multicolumn{2}{c}{7} & \multicolumn{2}{c}{8} & \multicolumn{2}{c}{9} \\ \cline{2-10} 
                             & Total Tokens & 60        & 80        & 60        & 80        & 60        & 80        & 60        & 80        \\ \hline
\multirow{4}{*}{T=0}         & HASS         &    4.39   &   4.49   &  4.49 & 4.62     &   4.54   &   4.67   &  4.59   &  4.73   \\
& PosS-1       &  4.54   &  4.64  &  4.65  &  4.78   &  \textbf{4.74}  &  4.89   &   \textbf{4.79}  &   4.94  \\
                             & PosS-2       &   \textbf{4.55}   &   \textbf{4.67}   &  \textbf{4.68}    &  \textbf{4.81}   &  \textbf{4.74}  &  \textbf{4.90}    &  \textbf{4.79}  &   \textbf{4.96}   \\
                             & PosS-3       &   4.50   &  4.62   &   4.61    &   4.75   &  4.69   &   4.83   &  4.73    &   4.89  \\ \hline
\multirow{4}{*}{T=1}         & HASS         &    4.16   &   4.24   &  4.22    &    4.34     &   4.26  &  4.39   &  4.30  &   4.41  \\
                             & PosS-1       &  \textbf{4.28}   &    \textbf{4.37}  &  4.35   &  4.48   &   \textbf{4.44}  &   \textbf{4.58}  &  4.47   & 4.58  \\
                             & PosS-2       &  4.27 &  \textbf{4.37}   &  \textbf{4.37}  &  \textbf{4.53}   &   4.43    &  4.57   &   \textbf{4.48}    & \textbf{4.64}   \\
                             & PosS-3       &  \textbf{4.28}   &   4.35    &   4.30   &   4.49  &  4.40  &  4.53  &  4.43  & 4.53  \\
\bottomrule
\end{tabular}
}
\end{table}

\begin{table}[h]
\caption{Speed-up ratio under different hyperparameters. Experiments use Llama-3-8B-Instruct as the base model. We average the results on all six datasets. The largest number within each row is bolded to show the upper bound of each method.
}
\label{tab:hyper-speedup-llama3}
\centering
\resizebox{\textwidth}{!}{
\begin{tabular}{cc|cccccccc}
\toprule
\multirow{2}{*}{Temperature} & Depth        & \multicolumn{2}{c}{6} & \multicolumn{2}{c}{7} & \multicolumn{2}{c}{8} & \multicolumn{2}{c}{9} \\ \cline{2-10} 
                             & Total Tokens & 60        & 80        & 60        & 80        & 60        & 80        & 60        & 80        \\ \hline
\multirow{4}{*}{T=0}         & HASS         & \textbf{2.89x}     & 2.83x     & 2.84x     & 2.78x   & 2.76x  & 2.71x  & 2.67x    & 2.65x    \\
                             & PosS-1       & \textbf{2.94x}  & 2.90x & 2.90x   & 2.85x   & 2.83x    & 2.80x   & 2.76x    & 2.72x   \\
                             & PosS-2       & \textbf{2.98x}  & 2.92x  & 2.93x   & 2.87x   & 2.84x   & 2.81x    & 2.77x   & 2.74x  \\
                             & PosS-3       & \textbf{2.95x}   & 2.89x   & 2.89x  & 2.84x     & 2.83x    & 2.78x    & 2.73x    & 2.71x  \\ \hline
\multirow{4}{*}{T=1}         & HASS         & \textbf{2.63x}        & 2.54x       & 2.56x      & 2.50x    & 2.47x     & 2.44x      & 2.41x     & 2.35x      \\
& PosS-1       & \textbf{2.73x}    & 2.65x   & 2.66x    & 2.59x  & 2.60x   & 2.55x   & 2.53x  & 2.48x  \\
                             & PosS-2       & \textbf{2.66x}  & 2.60x  & 2.63x    & 2.57x    & 2.55x  & 2.51x  & 2.48x  & 2.45x    \\
                             & PosS-3       & \textbf{2.67x}  & 2.59x  & 2.60x    & 2.56x    & 2.55x   & 2.47x  & 2.48x    & 2.41x    \\
\bottomrule
\end{tabular}
}
\end{table}

\begin{table}[h]
\caption{Average acceptance length under different hyperparameters. Experiments use Llama-2-13B-chat as the base model. We average the results on all six datasets. The largest average acceptance length within each column is bolded.
}
\label{tab:hyper-acc}
\centering
\resizebox{\textwidth}{!}{
\begin{tabular}{cc|cccccccc}
\toprule
\multirow{2}{*}{Temperature} & Depth        & \multicolumn{2}{c}{6} & \multicolumn{2}{c}{7} & \multicolumn{2}{c}{8} & \multicolumn{2}{c}{9} \\ \cline{2-10} 
                             & Total Tokens & 60        & 80        & 60        & 80        & 60        & 80        & 60        & 80        \\ \hline
\multirow{4}{*}{T=0}         & HASS         & 4.68      & 5.20      & 5.32      & 5.45      & 5.46      & 5.62      & 5.57      & 5.75      \\
& PosS-1       &   5.09    &  5.20   &  5.34  &   5.48   &   5.52   &   5.66  &   5.63    &   5.79  \\
                             & PosS-2       & \textbf{5.13}      & \textbf{5.22}      & 5.36      & 5.49      & 5.53      & 5.68      & 5.65      & 5.82      \\
                             & PosS-3       & \textbf{5.13}      & 5.21      & \textbf{5.37}      & \textbf{5.51}      & \textbf{5.55}      & \textbf{5.70}      & \textbf{5.66}      & \textbf{5.83}      \\ \hline
\multirow{4}{*}{T=1}         & HASS         & \textbf{5.00}         & 5.06         & 5.14      & 5.29         & 5.24       & 5.45       & 5.35     & 5.52     \\
                             & PosS-1       &   4.99    &   \textbf{5.11}    &   \textbf{5.24}    &   5.31   &   \textbf{5.34}   &   5.49    &   5.43   &  5.52   \\
                             & PosS-2       & 4.96      & \textbf{5.11}      & 5.16      & \textbf{5.32}      & 5.30      & 5.49      & \textbf{5.44}      & 5.61      \\
                             & PosS-3       & 4.99      & \textbf{5.11}      & 5.21      & 5.31      & 5.33      & \textbf{5.50}      & 5.43      & \textbf{5.62}    \\
\bottomrule
\end{tabular}
}
\end{table}

\begin{table}[h]
\caption{Speed-up ratio under different hyperparameters. Experiments use Llama-2-13B-chat as the base model. We average the results on all six datasets. The largest number within each row is bolded to show the upper bound of each method.
}
\label{tab:hyper-speedup}
\centering
\resizebox{\textwidth}{!}{
\begin{tabular}{cc|cccccccc}
\toprule
\multirow{2}{*}{Temperature} & Depth        & \multicolumn{2}{c}{6} & \multicolumn{2}{c}{7} & \multicolumn{2}{c}{8} & \multicolumn{2}{c}{9} \\ \cline{2-10} 
                             & Total Tokens & 60        & 80        & 60        & 80        & 60        & 80        & 60        & 80        \\ \hline
\multirow{4}{*}{T=0}         & HASS         & 3.28x     & 3.02x     & \textbf{3.33x}     & 3.08x     & 3.31x     & 3.09x     & 3.28x     & 3.09x     \\
                             & PosS-1       & 3.16x     & 2.93x     & \textbf{3.21x}     & 3.08x     & \textbf{3.21x}     & 3.09x     & 3.20x     & 3.09x     \\
                             & PosS-2       & 3.26x     & 3.00x     & 3.30x     & 3.06x     & \textbf{3.31x}     & 3.09x     & 3.27x     & 3.07x     \\
                             & PosS-3       & 3.29x     & 3.00x     & 3.34x     & 3.09x     & \textbf{3.35x}     & 3.11x     & 3.30x     & 3.10x     \\ \hline
\multirow{4}{*}{T=1}         & HASS         & 3.24x         & 2.94x         & \textbf{3.25x}         & 3.00x         & 3.20x         & 3.01x        & 3.18x         & 2.98x         \\
& PosS-1       & 3.13x     & 2.93x     & \textbf{3.17x}     & 2.95x     & 3.14x     & 2.97x     & 3.10x     & 2.92x     \\
                             & PosS-2       & 3.17x     & 2.94x     & \textbf{3.19x}     & 2.98x     & 3.18x     & 2.99x     & 3.17x     & 2.98x     \\
                             & PosS-3       & 3.24x     & 2.97x     & \textbf{3.26x}     & 3.00x     & \textbf{3.26x}     & 3.02x     & 3.18x     & 3.01x    \\
\bottomrule
\end{tabular}
}
\end{table}

\section{Extra Memory Usage During Inference}
\label{appendix:memory-usage}
\begin{figure}
    \centering
    \includegraphics[width=0.8\linewidth]{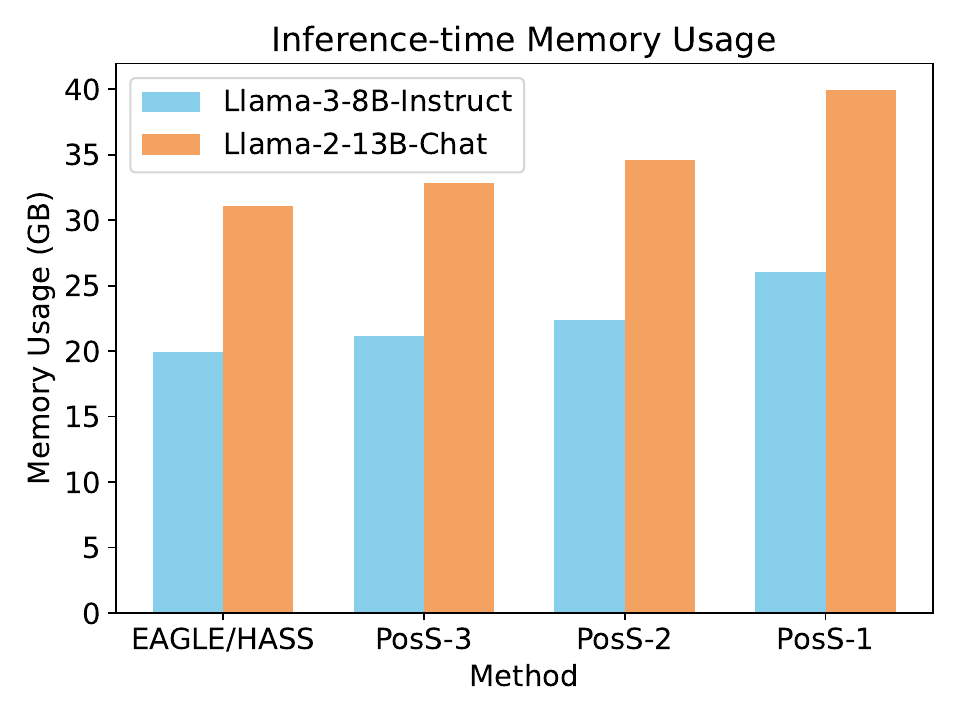}
    \caption{The Inference-time GPU memory usage of different speculative decoding methods. The memory usage is measured on the MT-bench test dataset. \method~methods require slightly more GPU memory than EAGLE-2, the baseline method.}
    \label{fig:memory-usage}
\end{figure}

In Section~\ref{sec:extra_cos}, we admit the \method~method requires slight extra GPU memory usage. Figure~\ref{fig:memory-usage} visualizes the memory usage of all methods mentioned in this paper. Here, EAGLE-2 and HASS cost the same GPU memory, and they are de facto \method-$\infty$. From left to right, the draft layers in the methods are 1, 2, 3, and 6. In both target model settings, \method-3 and \method-2 increase few extra memory usage. \method-1, despite using 6 times draft layers than EAGLE-2, costs acceptable extra memory usage.


\newpage

\end{document}